\documentclass[letterpaper]{article} 
\usepackage{aaai23}  
\usepackage{times}  
\usepackage{helvet}  
\usepackage{courier}  
\usepackage[hyphens]{url}  
\usepackage{graphicx} 
\usepackage{natbib}  
\usepackage{caption} 
\usepackage{algorithm}
\usepackage{algorithmic}
\usepackage{booktabs}
\usepackage{microtype}

\usepackage{newfloat}
\usepackage{listings}
\DeclareCaptionStyle{ruled}{labelfont=normalfont,labelsep=colon,strut=off} 
\floatstyle{ruled}
\newfloat{listing}{tb}{lst}{}
\floatname{listing}{Listing}
\title{Integrating Biological Data into Autonomous Remote Sensing Systems for In Situ Imageomics: A Case Study for Kenyan Animal Behavior Sensing with Unmanned Aerial Vehicles (UAVs)}
\author{
    Jenna M. Kline \textsuperscript{\rm 1},
    Maksim Kholiavchenko \textsuperscript{\rm 2},
    Otto Brookes \textsuperscript{\rm 3},
    Tanya Berger-Wolf \textsuperscript{\rm 1},
    Charles V. Stewart \textsuperscript{\rm 2}, 
    Christopher Stewart \textsuperscript{\rm 1},
}
\affiliations{
    \textsuperscript{\rm 1}The Ohio State University \\

    \textsuperscript{\rm 2}Rensselaer Polytechnic Institute \\

    \textsuperscript{\rm 3}University of Bristol \\
    
    kline.377@buckeyemail.osu.edu
}

\usepackage{bibentry}

\begin{document}

\maketitle

\begin{abstract}
In situ imageomics leverages machine learning techniques to infer biological traits from images collected in the field, or in situ, to study individuals organisms, groups of wildlife, and whole ecosystems. Such datasets provide real-time social and environmental context to inferred biological traits, which can enable new, data-driven conservation and ecosystem management. The development of machine learning techniques to extract biological traits from images are impeded by the volume and quality data required to train these models. Autonomous, unmanned aerial vehicles (UAVs), are well suited to collect in situ imageomics data as they can traverse remote terrain quickly to collect large volumes of data with greater consistency and reliability compared to manually piloted UAV missions. However, little guidance exists on optimizing autonomous UAV missions for the purposes of remote sensing for conservation and biodiversity monitoring. The UAV video dataset curated by \textit{KABR: In-Situ Dataset for Kenyan Animal Behavior Recognition from Drone Videos} \cite{kabr} required three weeks to collect, a time-consuming and expensive endeavor. Our analysis of \textit{KABR} revealed that a third of the videos gathered were unusable for the purposes of inferring wildlife behavior. We analyzed the flight telemetry data from portions of UAV videos that were usable for inferring wildlife behavior, and demonstrate how these insights can be integrated into an autonomous remote sensing system to track wildlife in real time. Our autonomous remote sensing system optimizes the UAV's actions to increase the yield of usable data, and matches the flight path of an expert pilot with an 87\% accuracy rate, representing an 18.2\% improvement in accuracy over previously proposed methods. 
\end{abstract}

\section{Introduction}
\begin{figure}[t]
    \centering
    \includegraphics[width=0.9\columnwidth]{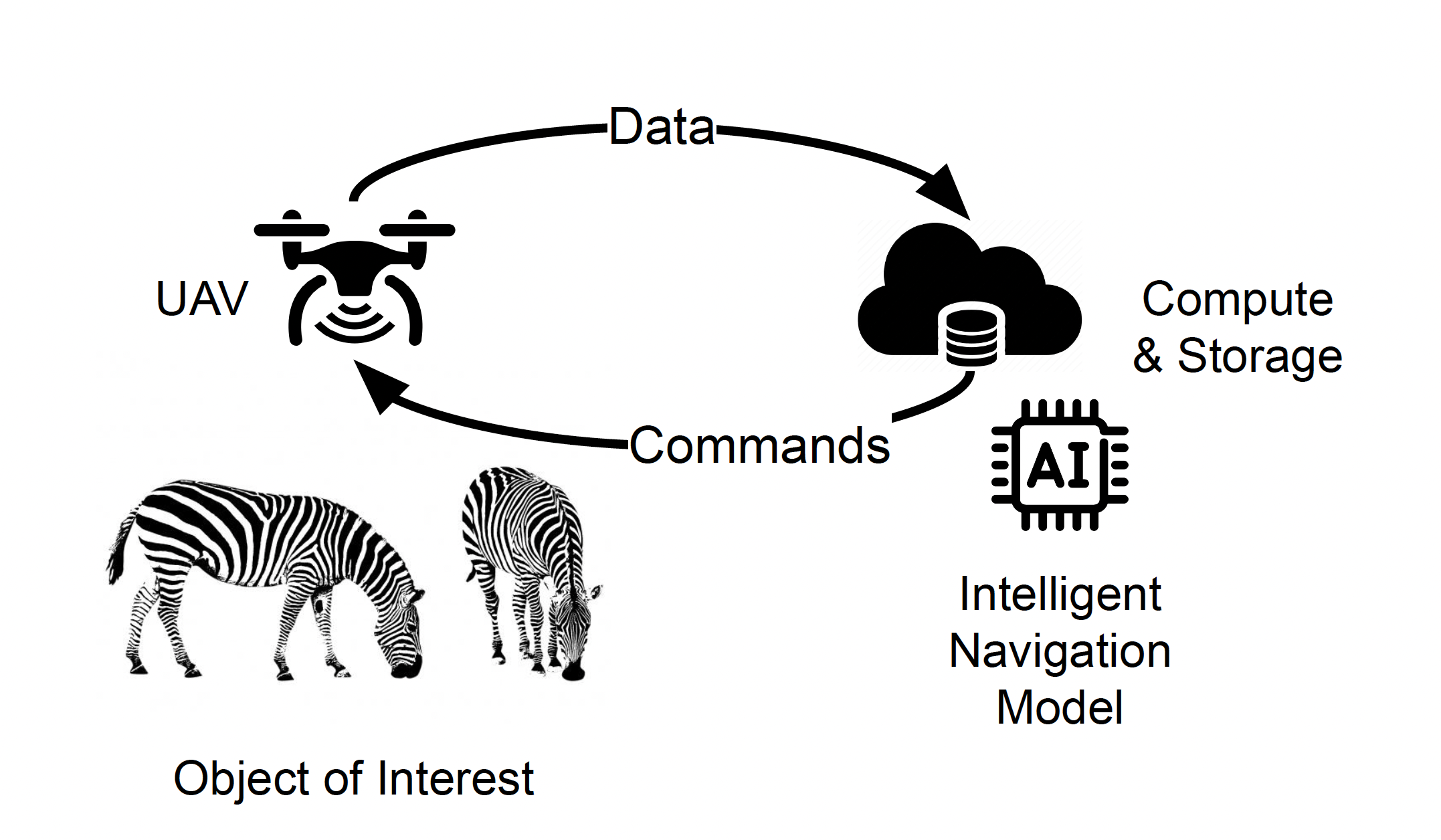} 
    \caption{Autonomous Remote Sensing System for In Situ Imageomics}
    \label{system}
    \end{figure}

Imageomics is a new scientific field that aims to understand and analyze biological traits of organisms, species, populations, and ecosystems from images \cite{imageomics}.  In situ imageomics focuses on studying biological organisms and systems within the context of their natural environment. Sources of such datasets include photos and videos captured by experts using camera traps \cite{Brookes_Mirmehdi_Kühl_Burghardt_2023, iot_camera_trap} , UAVs \cite {Koger_Deshpande_Kerby_Graving_Costelloe_Couzin, Andrew_Greatwood_Burghardt_2019}, and citizen scientists, such as iNaturalist \cite{inaturalist}. As wildlife respond to changing environmental conditions due to climate change, new approaches to remote sensing data collection and analysis are required to develop data-driven approaches to conservation and ecosystem management. Technological advancements in remote sensing data collection technology, namely camera traps \cite{Brookes_Mirmehdi_Kühl_Burghardt_2023, camera_traps_2019} and UAVs \cite{kabr, Koger_Deshpande_Kerby_Graving_Costelloe_Couzin}, have made it easier to gather large volumes of biological data more quickly than traditional manual techniques. With an imageomics approach, biological insights can be extracted from these datasets at scale using machine learning techniques. However, these models require massive volumes of quality data far exceeding what it typically captured for biological studies. This is particularly vital for models built to accomplish specific tasks, such as inferring behavior with the environmental and social context, which is difficult to infer from camera traps or data gathered by citizen scientists.

UAVs are well suited to gather the volume of data required to train such machine learning models for specific tasks. UAVs can dynamically track wildlife to capture fine-grained details, such as behavior,  while traversing  remote terrain quickly with little disturbance to the surrounding landscape. Experts pilots are required to conduct UAV missions tailored to the specific species and region. The ability to conduct such specialized missions with  autonomous UAVs would significantly reduce the barriers to conduct this research. Previous works have demonstrated that autonomous UAV missions are 3x cheaper, safer, more consistent, and reliable than manual flights \cite{Boubin_Chumley_Stewart_Khanal_2019, ACSOS23}. 

Existing autonomous UAVs methods are not optimized for in situ imageomics data collection. Autonomous UAVs may lose track of individuals over large distances, or fail to capture fine-grain details such as social dynamics. For both autonomous and manual approaches, crucial time may wasted on collecting and storing images with poor lighting and viewing angles. Such datasets require more storage and transport costs, and require more manual pre-processing before they can be analyzed. Our analysis of a three-day subset of the \textit{KABR} dataset \cite{kabr} revealed that only 2.2 hours of video was captured, and of that, only 65\% was usable for the mission objective of inferring wildlife behavior. UAVs have only recently become inexpensive enough for widespread use in biology applications, so little guidance exists on optimal strategies for maximizing the usable data yielded from such missions. 

In this work, we analyze the flight telemetry from portions of videos successfully used to extract behavior data in \textit{KABR}. We demonstrate how these insights can be integrated into a remote sensing system to improve the performance of autonomous UAVs to gather videos of wildlife behavior, with Kenyan animal behavior as a use-case. Figure \ref{system} illustrates such an autonomous remote sensing system for in situ imageomics. We show that dynamically adjusting the altitude and minimizing the movement of the UAV while ensuring that the wildlife are in view leads to higher usability rate of the collected videos.

\section{Methodology}
\begin{table*}[t]
\centering
\begin{tabular}{lll}
    \toprule
    \textbf{Data} & \textbf{Description} & \textbf{Source} \\
    \midrule
    Date \& Time & Date and time of video & Telemetry \& KABR \\
    Frame & Video frame (30 frames per second) & Telemetry \& KABR \\
    Behavior & Behavior of the animal & KABR \\
    Bounding Box Dimensions & Dimensions of the bounding boxes surrounding the animal in pixels & KABR \\ 
    Speed & Velocity of the UAV  in x,y,z direction (meters per second) & Telemetry \\
    Altitude & Altitude of UAV above ground (meters) &  Telemetry \\
    \bottomrule
\end{tabular}
\caption{Telemetry and KABR Dataset\citet{kabr}}
\label{data}
\end{table*}

We analyzed the flight telemetry data from portions of the UAV videos successfully used to extract behavior data from \textit{KABR: In-Situ Dataset for Kenyan Animal Behavior Recognition from Drone Videos} from \citet{kabr}. The videos were collected at the Mpala Research Centre in Laikipia, Kenya using a DJI Air 2S in 4K resolution over a period of three weeks in January 2023. The dataset includes three species: Plains zebra ({\it Equus quagga}), Grevy's zebra ({\it Equus grevyi}), and reticulated giraffes ({\it Giraffa reticulata}). The \textit{KABR} dataset includes the bounding box coordinates for each animal in the frame, the behavior annotations for each bounding box in each frame of the video, and the frame number and timestamp information. Since we were only interested in frames that contained usable behavior data, we excluded the subset of frames containing behavior annotations of ``Occluded'' ``Out of Frame'' or ``Out of Focus'' from our count of usable frames. Next, we obtained the telemetry data produced by the manually flown UAV  missions. The telemetry data for each mission began at the time the UAV  was powered on and launched and ended once the UAV was returned to the pilot and powered off. The telemetry data contains data gathered during the UAV flight which describes the status of the system, including altitude, speed, direction, and battery level. We used the frame number and timestamp information to reconcile the behavioral annotations with the corresponding telemetry data, as shown in Table \ref{data}. 
The telemetry data used in the analysis can be found here: https://huggingface.co/datasets/imageomics/KABR-telemetry
After reconciling the available telemetry and behavioral annotations, our final dataset consisted of 133 minutes of raw videos from 10 missions over three days. Of this final dataset, only  87 minutes contained video frames of suitable quality to both detect the animals of interest and determine their behavior, which is a 65\% usability rate. In other words, a third of the video data collected in the field could not be used to infer animal behavior using computer vision methods. We analyzed the telemetry data to identify what range of altitudes and bounding box values produced video that was suitable to be analyzed with machine learning computer vision methods. Specifically, we focused on computer visions methods for object detection with YOLO \cite{ultralytics_2023} to automatically localize the animals in the scene and automatic behavior detection proposed in \textit{KABR} to automatically extract behavior. Using these insights, we improved the autonomous navigation UAV model proposed in \citet{ACSOS23} to increase the yield of usable video data. We implemented the autonomous navigation model from \citet{ACSOS23} using the Yolov8n model \cite{ultralytics_2023} as the object detection model to localize the animals. 

\section{Results and Discussion}
\begin{table*}[t]
\centering
\begin{tabular}{lllll}
    \toprule
    & \textbf{Altitude (m)} & \textbf{Velocity (m/s)} & \textbf{Bounding Box Width (pixels)} & \textbf{Bounding Box Length (pixels)} \\
    \midrule
    Mean & 17.55 & 0.62 & 106.22 & 110.40 \\
    Standard Deviation & 7.80 & 1.31 & 61.65 & 69.01 \\
    Minimum & 4.40 & 0.00 & 11.00 & 14.00 \\
    25\% & 12.70 & 0.00 & 61.00 & 61.67 \\
    50\% & 14.90 & 0.00 & 92.00 & 96.67 \\
    75\% & 19.80 & 0.63 & 136.00 & 137.00 \\
    Maximum & 71.60 & 12.47 & 583.00 & 767.00 \\
    \bottomrule
\end{tabular}
\caption{Telemetry Data Analysis: Ranges of Optimal Values for UAV altitude, velocity, and animal bounding box sizes in pixels}
\label{telemetry}
\end{table*}

\begin{figure}[t]
    \centering
    \includegraphics[width=0.9\columnwidth]{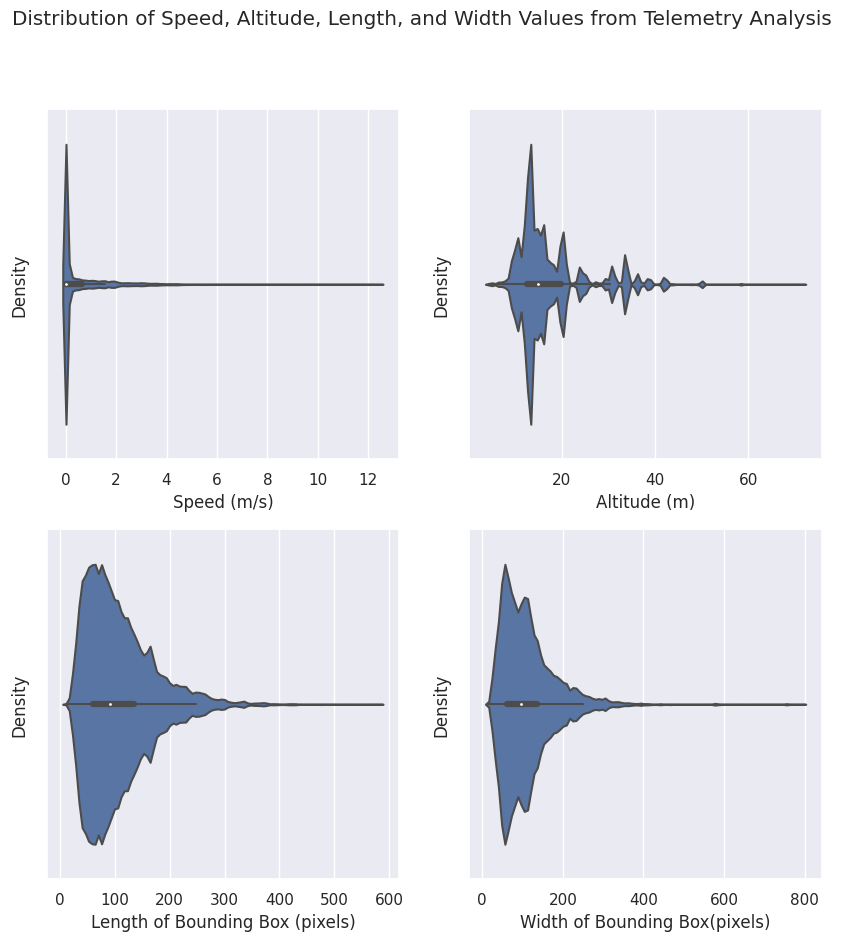}
    \caption{Distribution of altitude, velocity, and bounding box values associated with usable video data for wildlife behavior inference}
    \label{fig:violin}
\end{figure}

\begin{table*}[t]
\centering
\begin{tabular}{llll}
\toprule
  \textbf{Metric} & \textbf{Original Navigation Model} & \textbf{New \& Improved Navigation Model} & \textbf{Improvement} \\  
  \midrule
 \% of actions matching original flight & 68.8 & 87.0 & +18.2 \\
  F1 Score & 82.1 & 90.4 & +8.3 \\
  \bottomrule
\end{tabular}
\caption{Autonomous Navigation Model Performance Compared to Approach from~\cite{ACSOS23}}
\label{YOLO_nav}
\end{table*}

\begin{figure}[t]
\centering
\includegraphics[width=0.9\columnwidth]{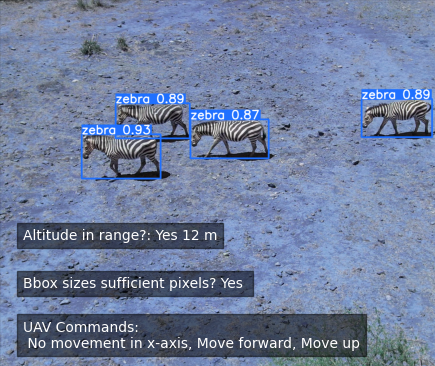} 
\caption{Example Autonomous Navigation Model Output}
\label{output}
\end{figure}

We used our telemetry analysis shown in Table~\ref{telemetry} and Figure \ref{fig:violin} to improve the autonomous navigation model for tracking wildlife proposed in \citet{ACSOS23}. This autonomous navigation model samples the video stream once every second, and utilizes a YOLO model \cite{ultralytics_2023} to automatically detect the animals in the frame. Once a zebra or giraffe is detected, the navigation model uses the bounding box dimensions generated by YOLO to determine the commands to send to the UAV. For situations when multiple animals are in the frame, this approach tracks the group by using the bounding box values to calculate the herd centroid. If this value strays sufficiently far from the centroid of the UAV's camera, the navigating model sends commands to the UAV to adjust its position to return the herd centroid to the center of the camera's view. One weakness of this autonomous navigation approach is that it is constrained to generating commands in x-y plane only (i.e. `fly left/right' and `fly forward/backward'), since this approach assumes a constant altitude. 

Our approach adds the ability to generate commands in z-direction (altitude), as well as the x and y directions, by integrating the insights from the telemetry analysis. This new and improved autonomous navigation model includes a range of preferred altitudes to tune the UAV's flight path. We also improve the autonomous navigation model by minimizing the movement of the UAV while navigating the UAV close enough for the bounding boxes to measure at approximately $100\times100$~pixels. This produces a 18.2\% increase in accuracy and a 8.3\% increase in the F1 score, as shown in Table \ref{YOLO_nav}. Figure \ref{output} shows the output of the autonomous navigation model proposed here. We report the F1 value, which combines the precision and recall scores of the model's predicted commands to the UAV. For this application, `positive' values means that the UAV is directed to move, while `negative' values means that the UAV is directed to hover in place. An example of a `true positive' result would be the model correctly predicting that the next command sent to the UAV should `fly left 3 meters'. An example of a `false positive' result would be the model incorrectly predicting that the next command sent to the UAV should `fly left 3 meters', when the correct decision was directing the UAV to hover in place, as determined by the expert pilot executing the manual mission. The accuracy values reported in Table~\ref{output} simply measures the number of predicted commands that match the original flight path conducted by the expert pilot. The telemetry dataset is imbalanced in favor of `positive' values, so the F1 score is a better measure of performance compared to accuracy.

\begin{figure}[t]
    \centering
    \includegraphics[width=0.9\columnwidth]{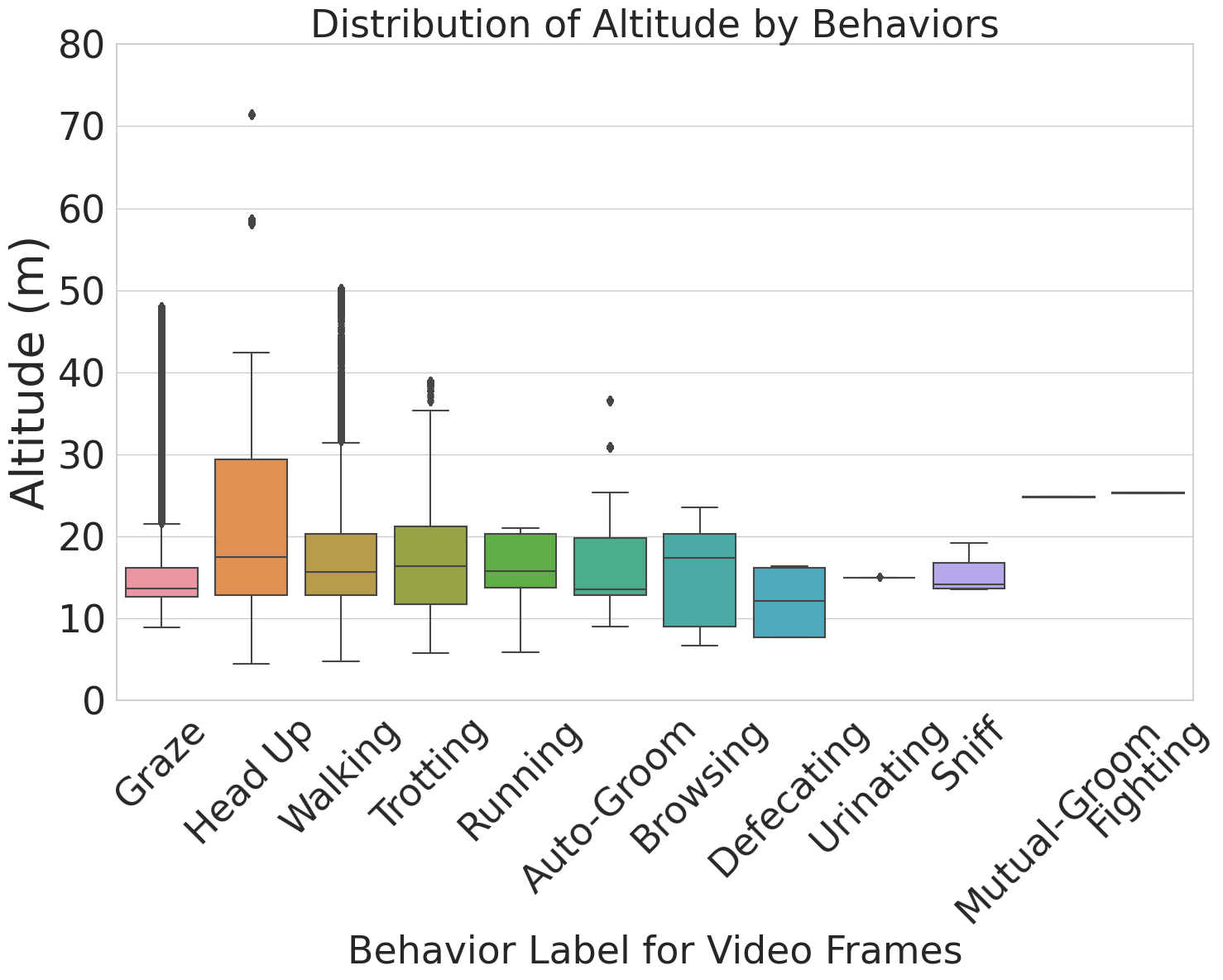} 
    \caption{Behavior Distribution by Altitude (behavior labels ordered from most to least common, left to right)}
    \label{behavioralt}
    \end{figure}

As shown in Table \ref{telemetry}, the mean altitude of the usable behavior video frames is 17.55 meters with a standard deviation of 7.8 meters. The maximum altitude is 71.60 meters, however the 75th percentile is 19.8 meters, so this maximum is an outlier.  Figure \ref{behavioralt} shows the relationship between behavior and altitude, where behaviors are ordered by how often they occur in the \textit{KABR} dataset. The majority of the behavior samples were collected below 30 meters, with rare behaviors such as `urinating' and `defecating' captured below 20 meters. These results suggest that behavior is best captured between 10 and 30 meters.

The mean velocity of usable video frames is 0.62 meters per second, with a standard deviation of 1.31 meters per second.  The distribution of velocity values is shown in Table~\ref{telemetry} and Figure~\ref{fig:violin}. For context, 0.62 meters per second is approximately 1.4 miles per hour.  The 75th percentile is only 0.63 meters per second, while the maximum velocity is 12.47 meters per second and the lower percentiles are 0 meters per second (i.e. hovering), suggesting the maximum value may be skewed by the UAV approaching the wildlife.  These results suggest that behavior is best captured by minimizing the UAV's movement and hovering once it reaches a suitable altitude, enabling the generation of bounding boxes of the appropriate size for behavioral analysis. 

\begin{figure}[t]
\centering
\includegraphics[width=0.9\columnwidth]{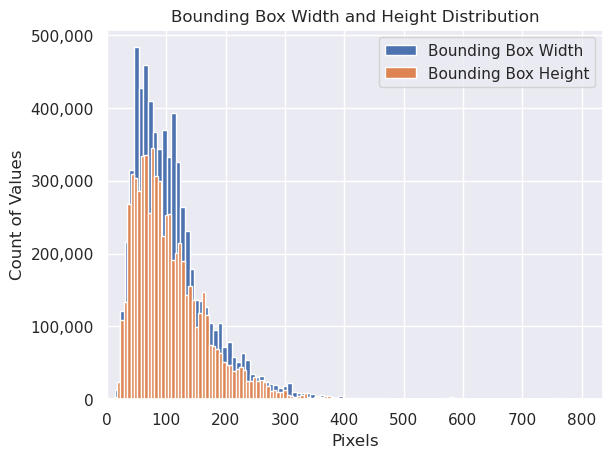} 
\caption{Bounding Box Dimension Distribution}
\label{bbox}
\end{figure}

The mean bounding box dimensions is approximately $106\times110$~pixels, with a standard deviation of $62\times69$~pixels. The distribution of bounding box dimensions is shown in Figure~\ref{bbox}. The bounding box size is related to the distance between the UAV and the wildlife. Using the bounding box size as an approximation for distance is an effective approach to tracking groups of free-ranging wildlife autonomously, as shown in  \citet{ACSOS23}. This approach eliminates the need for costly distance calculations, and can leverage the bounding box values produced by light-weight object detectors such as the YOLO nano model \cite{ultralytics_2023}.
    
\section{Conclusion \& Future Work}
We present guidelines for gathering in situ imageomics wildlife behavior video data with UAVs and demonstrate how these guidelines can be integrated into an autonomous UAV navigation model. The techniques proposed in this work can be applied to others' UAV wildlife video datasets to determine the optimal guidelines for additional species. These techniques can also be adapted to increase of the yield of usable UAV data beyond wildlife behavior studies; for example, individual identification of wildlife, such as those implemented by Wildbook \cite{bergerwolf2017wildbook}.   Future work will explore integrating behavior-adaptive flight data into autonomous UAV navigation models. This approach can minimize the disturbance from UAV data collection by using a behavior recognition model to determine if animals are exhibiting signs of reacting to the UAV's presence, such as vigilance behaviors or running. Reinforcement learning approaches can also be used to balance competing metrics, such as the range of altitudes and minimum desired bounding box size. These approaches should be validated on free ranging wildlife to confirm the proposed approaches produce data suitable for analysis with computer vision techniques for imageomics applications.
    
    
\section{Acknowledgments}
This work was supported by NSF Awards No. 2118240 and 2112606. Data was collected in accordance with Kenya’s National
Commission for Science, Technology \& Innovation research license NACOSTI/P/22/18214 and Princeton University’s Institutional Animal Care and Use Committee 1835F.

\bigskip

\bibliography{aaai23}

\end{document}